\documentclass[conference]{IEEEtran}
\ifCLASSINFOpdf
\else
\fi
\usepackage{graphicx}
\usepackage{subfigure}

\usepackage[numbers,sort&compress]{natbib}

\IEEEoverridecommandlockouts

\begin{document}

%
\title{Sentiment Analysis based on User Tag for Traditional Chinese Medicine in Weibo}

\author{\IEEEauthorblockN{Junhui Shen}
\IEEEauthorblockA{Information Center\\
Beijing Univercity of Chinese Medicine\\
Beijing, China\\
Email: shen-junhui@163.com}
\and
\IEEEauthorblockN{Peiyan Zhu, Rui Fan}
\IEEEauthorblockA{State Key Lab of Software Development Environment\\
Beihang University \\
Beijing, China 
}

}


%


\maketitle

\begin{abstract}
With the acceptance of Western culture and science, Traditional Chinese Medicine (TCM) has become a controversial issue in China. So, it's important to study the public's sentiment and opinion on TCM. The rapid development of online social network, such as twitter, make it convenient and efficient to sample hundreds of millions of people for the aforementioned sentiment study. To the best of our knowledge, the present work is the first attempt that applies sentiment analysis to the domain of TCM on Sina Weibo (a twitter-like microblogging service in China). In our work, firstly we collect tweets topic about TCM from Sina Weibo, and label the tweets as supporting TCM and opposing TCM automatically based on user tag. Then, a support vector machine classifier has been built to predict the sentiment of TCM tweets without labels. Finally, we present a method to adjust the classifier result. The performance of F-measure attained with our method is 97$\%$. 
\end{abstract}


%
\IEEEpeerreviewmaketitle

\section{Introduction}
With the rapid acceptance of Western culture and science from the beginning of the 20th century, Traditional Chinese Medicine (TCM) was seriously thrown into doubts in terms of its scientific foundation.  When such kind of debate was reviewed in respective of debaters’ sentiment towards TCM, it can be seen that two types of sentiment are dominating: one school thinks that TCM are not proved by scientific experiment, so it is pseudo-science and should be abolished, while the other school believes TCM is effective in treating many diseases and therefore TCM is essentially a kind of science. 

Microblogging today has become a very popular communication tool among internet users.  In China, Sina Weibo(http://www.weibo.com), a Twitter-like microblogging service launched in 2009, has accumulated more than 500 million users in less than four years, leading to it's most important role in the social media marketing platform. Every second, approximately more than 1000 Chinese tweets are posted in Weibo. It is imaginable that the debate surrounding TCM spread into cyber-space in unbelievable speed. 

So far, although many researches have been conducted on sentiment classification, there is very little such work done on Traditional Chinese Medicine. To the best of our knowledge, the present work is the first attempt that applies sentiment analysis to the domain  of TCM on Sina Weibo. In our work, main contents are: collecting corpus and dictionary resources, labeling data automatically based on user tag, building an SVM classifier to predict the sentiment of TCM tweets and presenting a method to adjust the classifier result. The performance of F-measure attained with our method is 0.97$\%$. 

The rest of the paper is organized as follows: In section II presents previous works on sentiment analysis and their application for microblogging.  In Section III data collecting and labeling are discussed. Feature selection and learning method are described in Section IV. In Section V experimental results are presented. Finally, conclusion and future directions of research are discussed in section VI.

 
\section{Related Work}
Sentiment classification has been investigated in different domains such as movie reviews, product reviews and customer feedback reviews. The main researches have fallen into two categories. The first is  machine learning techniques, which attempts to train a sentiment classifier based on occurrence frequencies of the various words in the documents. The other approach is semantic oriented, which classifies words into two classes, such as "positive" or "negative", and then counts an overall positive/negative score for the text.  A very broad overview of the existing work was presented in \cite{{pang2008opinion},{liu2012sentiment}}. In their survey, the authors describe existing techniques and approaches for sentiment analysis. 

With the popularization of microblog and online social networks, such as Twitter and Weibo, sentiment analysis become a field of interest to many researches. Some of the early and recent results on sentiment of Twitter data have been presented in \cite{{go2009twitter},{agarwal2011sentiment},{bermingham2010classifying}}. Go et al.\cite{{go2009twitter}} used distant learning to acquire sentiment data. They used tweets ending in positive emoticons like ":)",  ":-)" as positive and negative emoticons like ":(",  ":-(" as negative. They built models using Naïve Bayes, MaxEnt and Support Vector Machines (SVM), and they reported that SVM outperforms other classifiers. In terms of feature space, they tried unigram, bigram model in conjunction with parts-of-speech (POS) features. They noted that the unigram model outperforms all other models. Specifically, bigrams and POS features do not help. So, in this paper, we use the SVM classifier with a unigram model. In China, some of the research on sentiment of Weibo data are conducted by Zhao et al. \cite{{zhao2012moodlens}} and He\cite{{he2013sentiment}} etc. Zhao et al. \cite{{zhao2012moodlens}} built a system for sentiment analysis of Chinese tweets in Weibo. It employed the emoticons for the generation of sentiment labels for tweets, and built an incremental learning Naïve Bayes classifier for the categorization of four types of sentiments: angry, disgusting, joyful and sad. 

About the subject of TCM in Weibo, however, there is very little investigation conducted on sentiment classification.

\section{Data collecting and labeling }
In this section, we discuss the collecting and preprocessing of tweet topic on TCM. For each tweet in our corpus, we convert it into a sequence of words. 
\subsection{Corpus Collection based on User Tag}
In China, Sina Weibo is one of China's most important social networking channels, and is the Chinese counterpart to Twitter. As with Twitter, Weibo users are allowed to post real-time messages, called tweets. Tweets are short messages, restricted to 140 characters in length.

There are some prominent differences between Twitter and Weibo. For example, user can freely tag himself/herself to indicate his/her interests and characteristics in Weibo. Of course, Tagging is not mandatory in Weibo where users can tag up to 10 keywords. 

In January 2014, we searched Weibo users interested in TCM by user tag.  If someone has more than one user tag included in our search keywords list, he/she would be duplicated in our dataset. After filtering the duplicated users, we constructed a dataset including 48861 Weibo users, denoted as C. The user tags and the corresponding numbers of Weibo users are listed in Table 1. Among all tags, "Traditional Chinese Medicine" is used by 42608 users and occupies the dominating share of 87\%, "Medicine Material", "Acupuncture and Moxibustion" and "Massage" follow but none of them takes the share more than 8\%. It is not surprising "Traditional Chinese Medicine" is the main tag used because it is a wide concept, which often refers to not only TCM therapy but also including "Medicine Material", "Acupuncture and Moxibustion" and "Massage". 

\begin{table}[htbc]
\renewcommand{\arraystretch}{1.3}
\caption{User Tag and the Number of Weibo User  Correspondingly  }
\label{table:1}
\centering
\begin{tabular}{c|c|c}
\hline
\bfseries User Tag & \bfseries User Tag & \bfseries the Counts \\
(Original Text) & (English Translation) & of Weibo Users\\
\hline 
中医  & Traditional Chinese Medicine & 42608 \\
\hline
中药 & Medicine Material & 3827 \\
\hline
针灸 & Acupuncture and Moxibustion & 3236 \\
\hline
推拿 & Massage & 2198 \\
\hline
艾灸 & Moxa-moxibustion & 763 \\
\hline
中草药 & Chinese Herb Medicine & 417 \\
\hline
针刺 & Acupuncture & 73 \\
\hline
针推 & Acupuncture and Massage & 67 \\
\hline
中成药 & Chinese Patent Drug & 50 \\
\hline
\end{tabular}
\end{table}

Using the Application Programming Interfaces(APIs) provided by Weibo, we collected the tweets which were posted by the users in C.  Due to the limit of API, only the most recent  2000 tweets of each user posted can be obtained, we gathered  21,242,370 tweets totally.

The sentiment of a retweet is not always consistent with the tweet, especially when debating. For this reason, we split every tweet which has retweet and insert every retweet into our corpus. Sometimes, one post has more than one re-posting, so we have much more tweets after splitting.  Totally, we collected 43,012,068 tweets in our corpus, more than twice of original tweets count. 

\subsection{Two Dictionary Resources }
In this paper, we introduce two new resources for the preprocessing of Weibo data topic on TCM: custom dictionary and TCM terminology dictionary. We collect western medicine terminology, TCM terminology and popular vocabulary on the internet, totally 5307 words in the custom dictionary. It can be used as a helpful complement of  build-in dictionary of general tool for Chinese Word Segmentation. The TCM terminology dictionary collect 2715 TCM terminology words including Traditional Chinese Medicine, Chinese Patent Medicine, Chinese Herb Medicine and  acupuncture point etc. It can be used to filter the Weibo which topic is about TCM.
\subsection{Preprocessing of Data}
We pre-process all the tweets as follows:
1)	Translating the tweet to Chinese Simplified if it is written by Chinese Traditional;
2)	Filtering URL links (e.g. http://example.com ), Weibo user names (e.g. @shen –with symbol @ indicating a user name), Weibo special words ( e.g. reply ), and emoticons from tweets; 3)Segmenting Chinese Word (with the ICTCLAS tool \cite{{zhang2003hhmm}} and the custom dictionary as introduced in subsection B ) to generate a sequence of words; 4) Removing stop words ( such as "oh" ) from the bag of words; 5) Filtering advertisements by key words (such as "sale"). 
\subsection{Filtering Chinese Medicine Tweets}
However, the topics of tweets posted by the users interested in TCM are diverse and not only concerning to TCM. Therefore, in our study, we should screen out the tweets in which the real topic is not about TCM.

In our approach, we filter the tweets topic on TCM with the TCM terminology dictionary (introduced in subsection B).  Usually, a tweet topic on TCM contains more than one key word, so we filter the tweets including at least two different key words of TCM strictly. After filtering, there remain in our corpus 1,650,497 tweets in which the real topic is about TCM. 
\subsection{Labeling the Data}
When we label the sentiment of tweet, our approach is based on the basic principle: the user is prone to have consistent opinions for a certain topic due to the principle of consistency \cite{{deng2014exploring}}. It means that if the user's opinion is for TCM, the sentiment of all the tweets he/she posted is for TCM. In contrast, if the user's opinion is against TCM, the sentiment of all the tweets he/she posted is against TCM. 

In our approach, we acquire user's opinion about TCM by the user tag. The keyword which is used as user tag is defined by the user.  Consequently, the user tags could be different even if the sentiment to TCM is same. The user tags which are used to label the sentiment are listed in Table 2. Only the user tags which have been quoted by more than 10 users are included in the table. As a result, 1866 Weibo users are labeled as supporting TCM, while 290 Weibo users are labeled as opposing TCM. The rest are not labeled because we can't obtain obvious sentiment orientation from his/her user tags.

\begin{table}[t]
\renewcommand{\arraystretch}{1.3}
\caption{User Tag of different sentiment and corresponding Weibo user counts }
\label{table:2}
\centering
\begin{tabular}{c|c|c|c}
\hline
\bfseries Sentiment & \bfseries User Tag & \bfseries User Tag & \bfseries User\\
& (Original Text) & (English Translation) & Counts\\
\hline 
Supporting TCM 
& 中医爱好 & Love TCM & 972 \\
& 爱中医 & Love TCM & 239 \\
& 中医师 & Doctor of TCM & 230 \\
& 喜欢中医 & Love TCM & 85 \\
& 中医粉 & TCM Follower & 55 \\
& 中医控 & TCM Follower & 52 \\
& 中药师 & Pharmacist of TCM & 51 \\
& 针灸师 & Acupuncturist & 42 \\
& 中医养生爱好 & Regimen of TCM & 29 \\
& 推拿师 & Masseur & 28 \\
& 中医达人 & TCM Master & 12 \\
\hline
Opposing TCM
& 反中医 & Oppose TCM & 191\\
& 中医黑 & Abominate TCM & 55\\
& 反对中医 & Oppose TCM & 28\\
\hline
\end{tabular}
\end{table}

Based on our basic principle, we label the sentiment of tweet according to the user's opinion on TCM. Finally, 40888 tweets are labeled as supporting TCM, and 6975 tweets are labeled as opposing TCM. Obviously, there is an imbalance but it is consistent with the real world.
The tweets labeled will be used as the training dataset in the next step of our research. 
\section{Methodology}
This section presents the methodology of sentiment classification system we use. First, feature selection method is used to pick out discriminating terms for training and classification. Then we use the machine learning method to build a sentiment classifier. Finally, we adjust the classification result based on the basic principle that a user keeps consistent opinions for a certain topic.
\subsection{Feature Selection}
A number of feature selection metrics have been explored in text categorization, i.e. chi-square (CHI), information gain (IG), correlation coefficient (CC) and odds ratios (OR). All these methods compute a score for each individual feature and then pick out a predefined size of feature set. In our approach, we use the chi-square feature selection method, one of the most effective methods in text categorization \cite{{yang1997comparative}}.

Chi-square measures the lack of independence between a term t and a category c$_i$ and can be compared to the chi-square distribution with one degree of freedom to judge extremeness. It is defined as:

\begin{displaymath}
\chi^2(t,c_i)=\frac{N[P(t,c_i)P(\bar{t},\overline{c}_i)-P(t,\overline{c}_i)P(\bar{t},c_i)]^2}{P(t)P(\bar{t})P(c_i)P(\overline{c}_i)}
\end{displaymath}
where N is the total number of documents.

\subsection{Machine Learning Method}
So far, most of the research on sentiment classification focused on training machine learning algorithms to classify reviews. Support vector machine (SVM) have been shown to be highly effective for traditional text categorization \cite{{pang2002thumbs}}. 

Based on the structural risk minimization principle from the computational learning theory, SVM seeks a decision surface to separate the training data points into two classes and makes decisions based on the support vectors that are selected as the only effective elements in the training set. 

Here we limit our discussion to linear SVM due to its popularity and high performance in text categorization \cite{{fan2008liblinear}}. 

The optimization of SVMs (dual form) is to minimize:
\begin{displaymath}
\vec{\alpha}^*=arg \ min\{-\sum_{i=1}^n \alpha_i + \sum_{i=1}^n \sum_{j=1}^n \alpha_i\alpha_jy_iy_j<\vec{x}_i,\vec{x}_j> \}
\end{displaymath}
\begin{displaymath}
Subject \ to: \sum_{i=1}^n \alpha_i y_i=0;\quad 0\leq \alpha_i\leq C
\end{displaymath}

For a tutorial on SVM and details of their formulation we refer the reader to Burges \cite{{burges1998tutorial}} and Cristiani \cite{{cristianini2000introduction}}. A detailed treatment of these models’ application to text classification can be found in Joachims \cite{{joachims2002learning}}.

\subsection{Adjusting Sentiment Classification Result}
Based on the basic principle that the same user should have consistent opinions for a certain topic, we adjust the sentiment classification result: assign majority sentiment label to all the tweets the same user posted.

Based on the sentiment classification result, the number of tweets which are judged as supporting TCM posted by one user can be obtained as C$_s$, and the number of tweets  posted by the same user which are judged as opposing TCM can be obtained as C$_o$. Then we define $\gamma$ as 
\begin{displaymath}
\gamma=\frac{max\{C_s,C_o\}}{C_s+C_o}
\end{displaymath}
where  $0.5\leq\gamma\leq1$. If $\gamma$ = 1, it means the sentiment of the user is consistent absolutely. If $\gamma$ = 0.5, it means Co is equal to Cs, then we don't need to adjust the sentiment classification result. When $0.5<\gamma<1$, we can adjust the classification result. 

\begin{table*}[t]
\renewcommand{\arraystretch}{1.3}
\caption{
\bf{Top 10 Key Words of Each Class}}
\centering
\begin{tabular}{c|c|c|c}
\hline
\bfseries Supporting TCM  & \bfseries Supporting TCM  & \bfseries Opposing TCM & \bfseries Opposing TCM\\
(Original Text) & (English Translation) & (Original Text) & (English Translation)\\
\hline 
中药 & Medicine Material & 中成药 & Chinese Patent Medicine \\
养生 & Health Preservation & 马兜铃酸 &  Aristolochic acid \\
国家 & State & 注射 & injection \\
 科学 & Science & 注射液 & injection \\
 中医药 & TCM & 方舟子 &  Zhouzi Fang \\
 中国 & China & 朱砂 & Cinnabar \\
 身体 & Body & 事件 & Events \\
 医生 & Doctor & 反对 & Oppose \\
 健康 & Health & 马兜铃 & Aristolochic \\
 治疗 & Cure & 龙胆泻肝丸 & Longdan Xiegan Wan\\
\hline
\end{tabular}

\begin{flushleft}
\end{flushleft}
\label{table:3}
\end{table*}

\section{Experiments and Results}
In our dataset, there are 1,650,497 tweets in which the topic focuses on TCM, including 40,888 tweets labeled as supporting TCM, and 6,975 tweets labeled as opposing TCM ( introduced in Section 4 ). Since it's imbalanced, we focused on not only the global performance, but also the performance of each class. Therefore, we choose the F1 to evaluate the classification system.

After applying CHI feature selection to tweets, for all our experiments we use Support Vector Machine and report 5-fold cross-validation test results. 

Pang \cite{{pang2002thumbs}} argued that feature presence binary value is more useful than feature frequency for the SVM classier. Therefore, we use binary value for each feature instead of feature frequency.

\subsection{The performance measure}
To evaluate the imbalanced classification system, we use the F$_1$ measure. This measure combines recall and precision in the following way:
\begin{displaymath}
Precision=\frac{number \ of \ correct \ positive   \ predictions}{number \ of \ positive \ predictions}
\end{displaymath}
\begin{displaymath}
Recall=\frac{number \ of \ correct \ positive \ predictions}{number \ of \ positive \ examples}
\end{displaymath}
\begin{displaymath}
F_1=\frac{2*Precision*Recall}{Recall+Precision}
\end{displaymath}

\subsection{Feature Selection Results}
The top 10 key-words of each class selected by the CHI method are listed in table 3. Among the proponents of TCM, it is not surprise that "Medicine Material", "Health Preservation", "Tradational Chinese Medicine" and "Body" are often used. The frequncy of "State" and "China" could be due to that Chinese government employed clear policy to support TCM. Among the opponents of TCM, "Aristolochic acid", "Cinnabar", "Longdan Xiegan Wan" and "injection" are popular words. This could be due to that all these terms are related to untoward effects so the opponents want to shake the scientific foundation of TCM.

Figure 1 shows the classification performance curves using the CHI feature selection method vs. feature number. The performance of classifier is above 90$\%$ stably and the performance increases  as number of features increases. It is found that the performance of TCM proponent classifier is slightly higher than the performance of the total classifier. It is notable that the performance of TCM opponent classifier increase significantly when number of features increases. The performance of each class is relatively stable when the number of features exceeds 3000. So, we fixed the number of features at 3000 in the following experiments. 

\begin{figure}[h!]
\centering
\includegraphics[width=5cm]{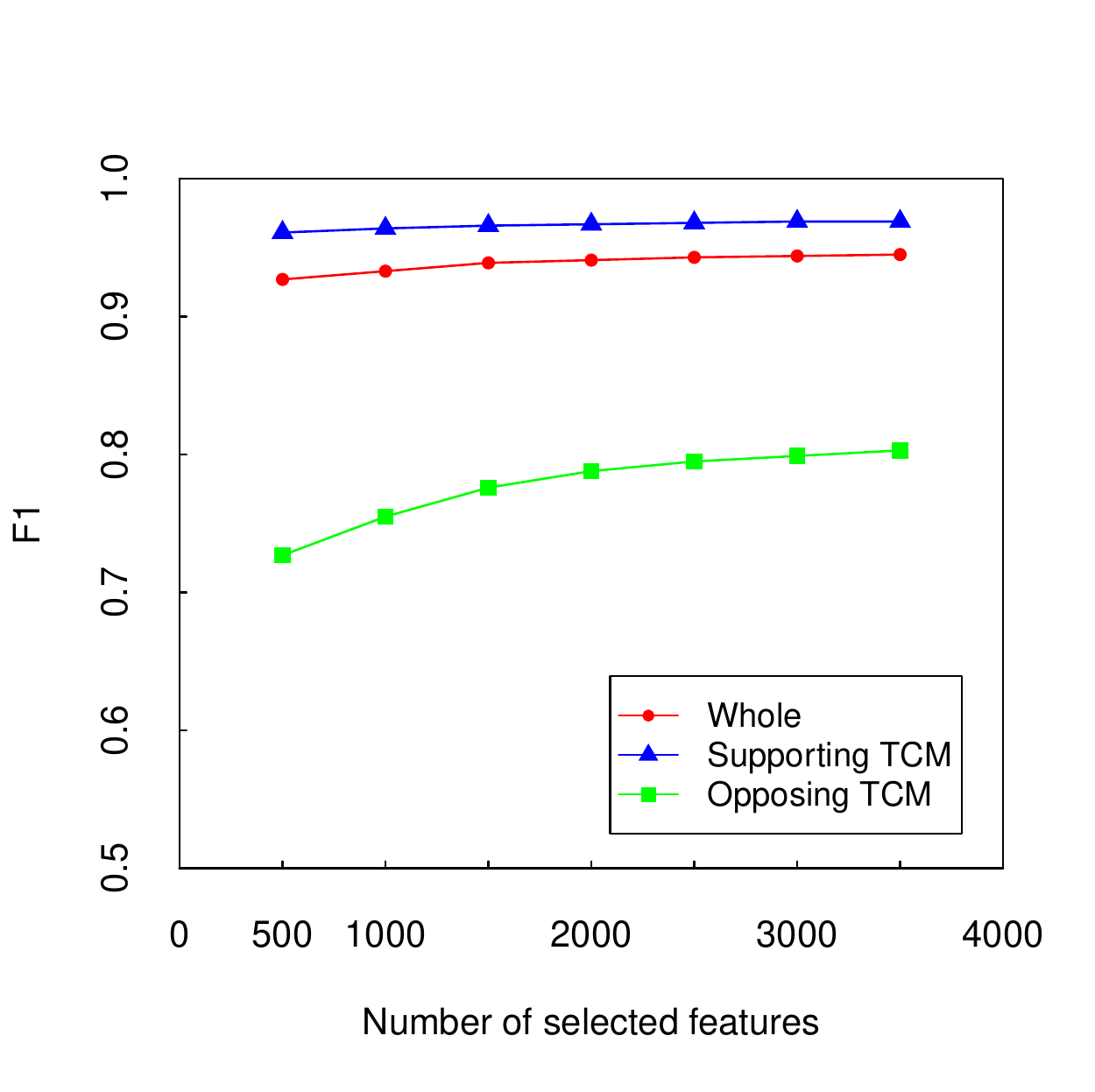}
\caption{
{ The Performance Curves of Each Class vs. Feature Number.} 
 }

\label{Figure_1}
\end{figure}

\begin{figure}[h!]
 \centering
  \includegraphics[width=5cm]{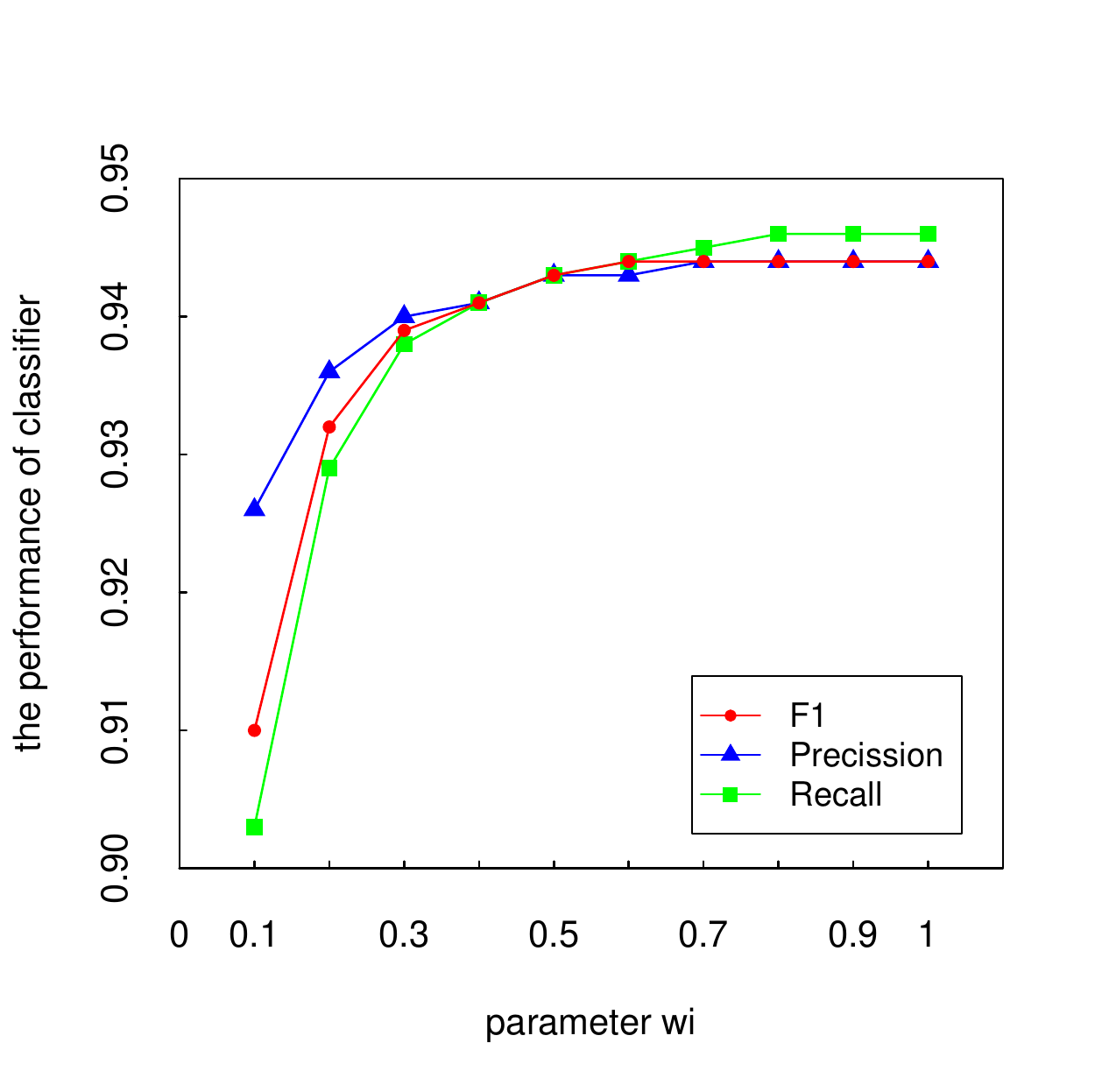}
 \caption{
 { The Performance Curves vs. Parameter wi} 
 }
 \label{Figure_2}
 \end{figure}

 \begin{figure}[t!]
 \centering
 \subfigure[PRECISION]{
  \includegraphics[width=4cm]{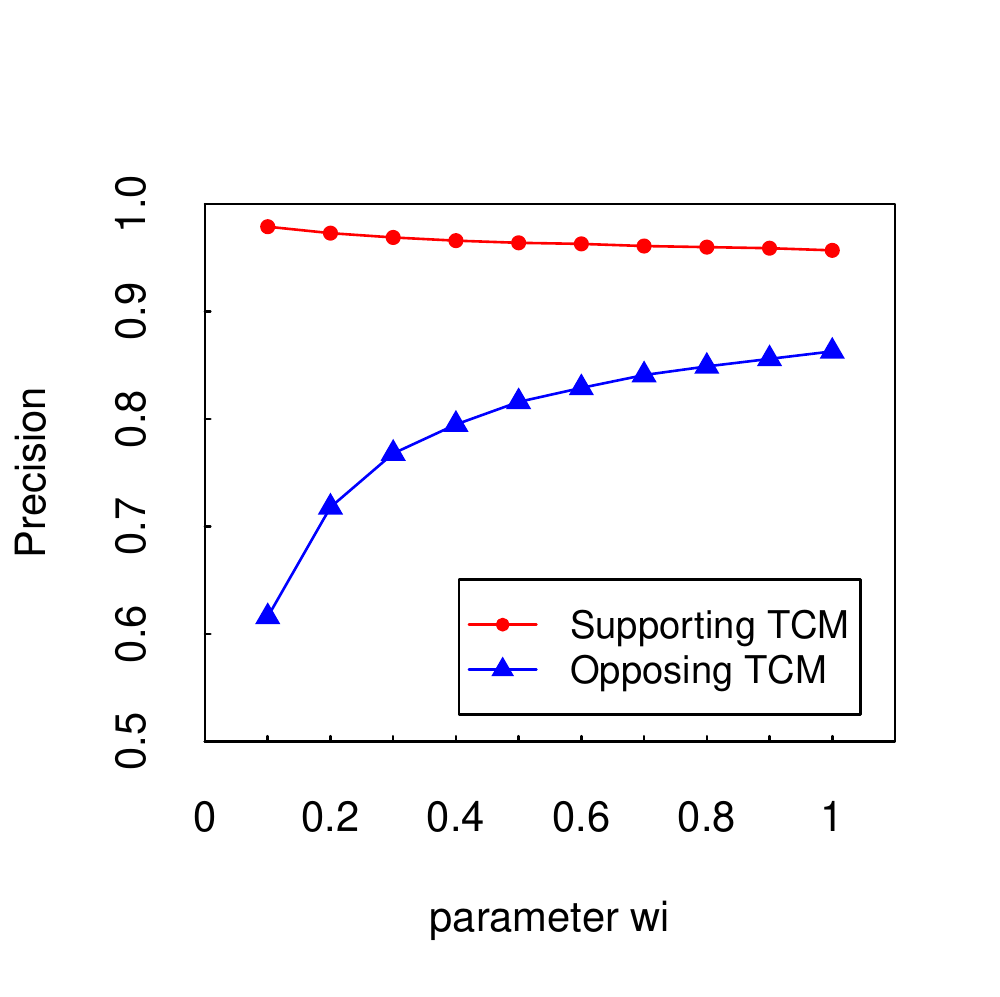}
  \label{}
  }
 \subfigure[RECALL]{
 \includegraphics[width=4cm]{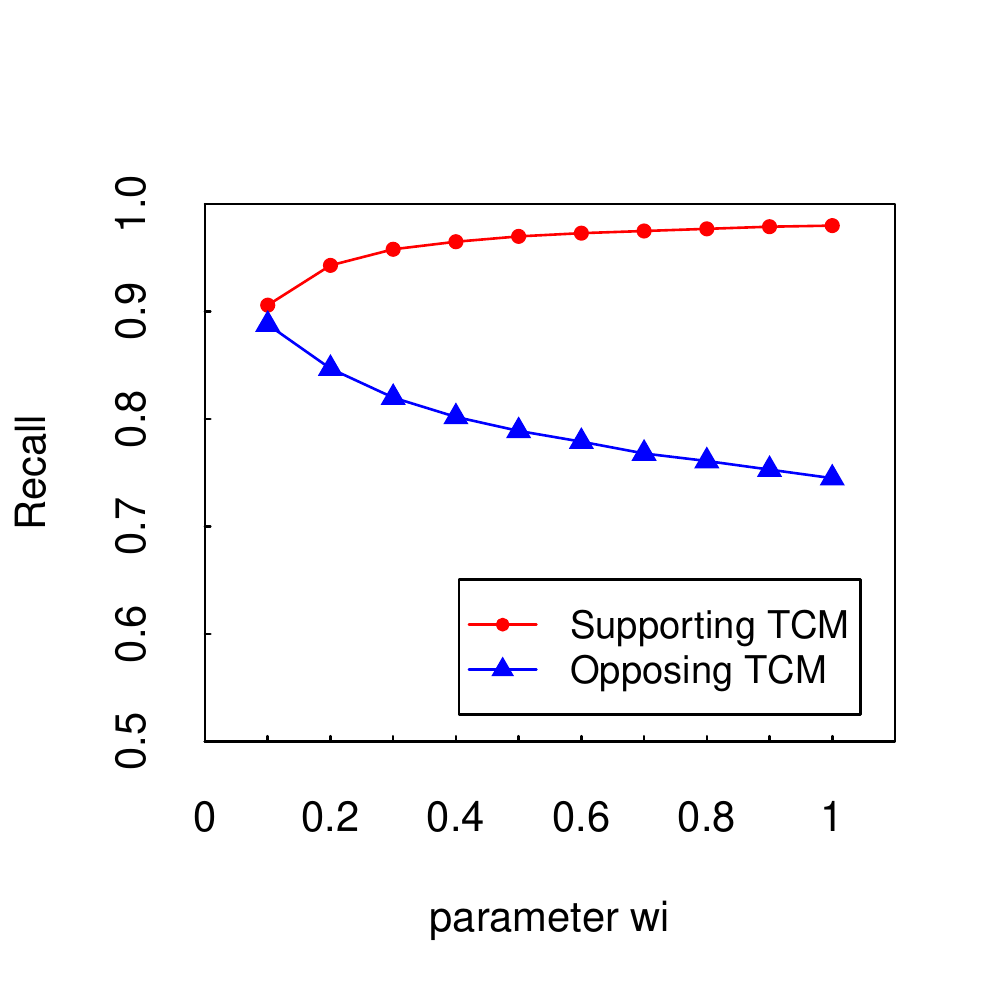}
 \label{}
 } 
  
 \caption{
 {\bf The Precision and Recall of Each Class vs. Parameter wi} 
 } 
 \label{Figure_3}
 \end{figure}

 \begin{figure}[t!]
 \centering
 \subfigure[SUPPORTING TCM]{
     \includegraphics[width=4cm]{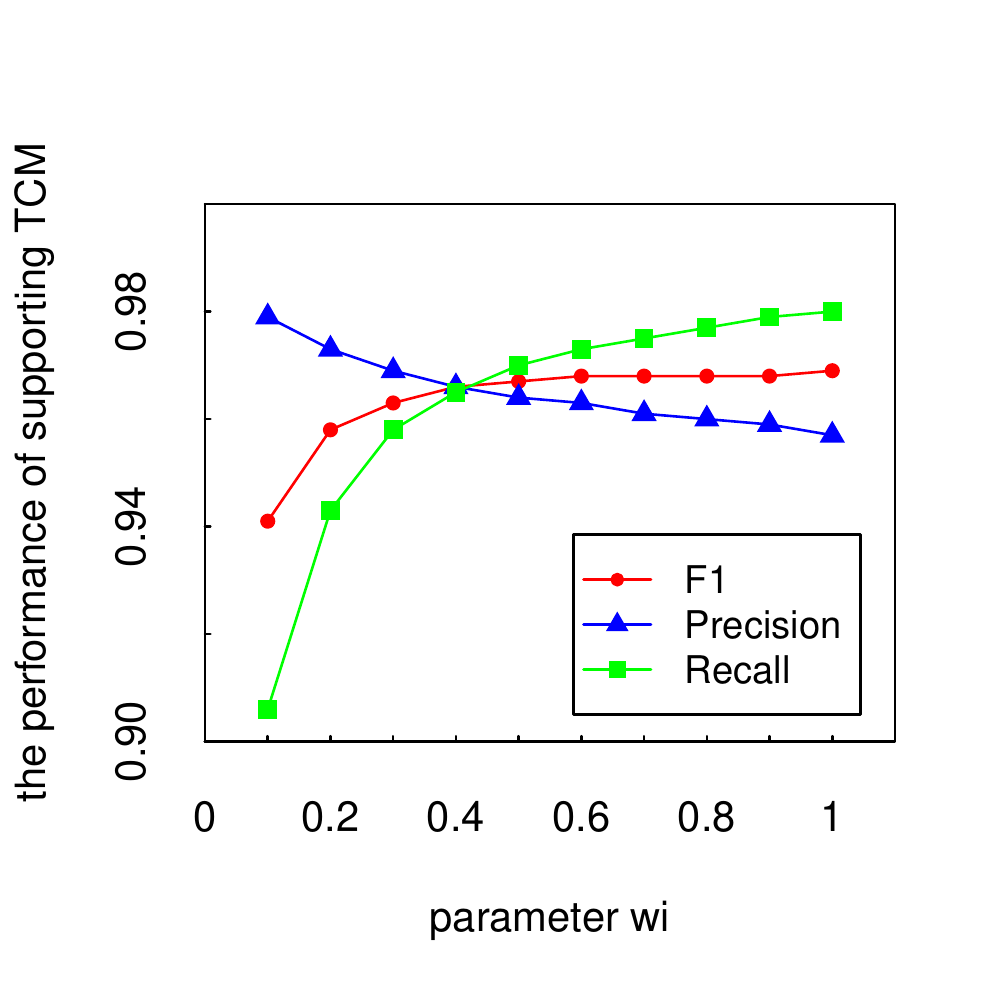}
     \label{}
     }
  \subfigure[OPPOSING TCM]{
     \includegraphics[width=4cm]{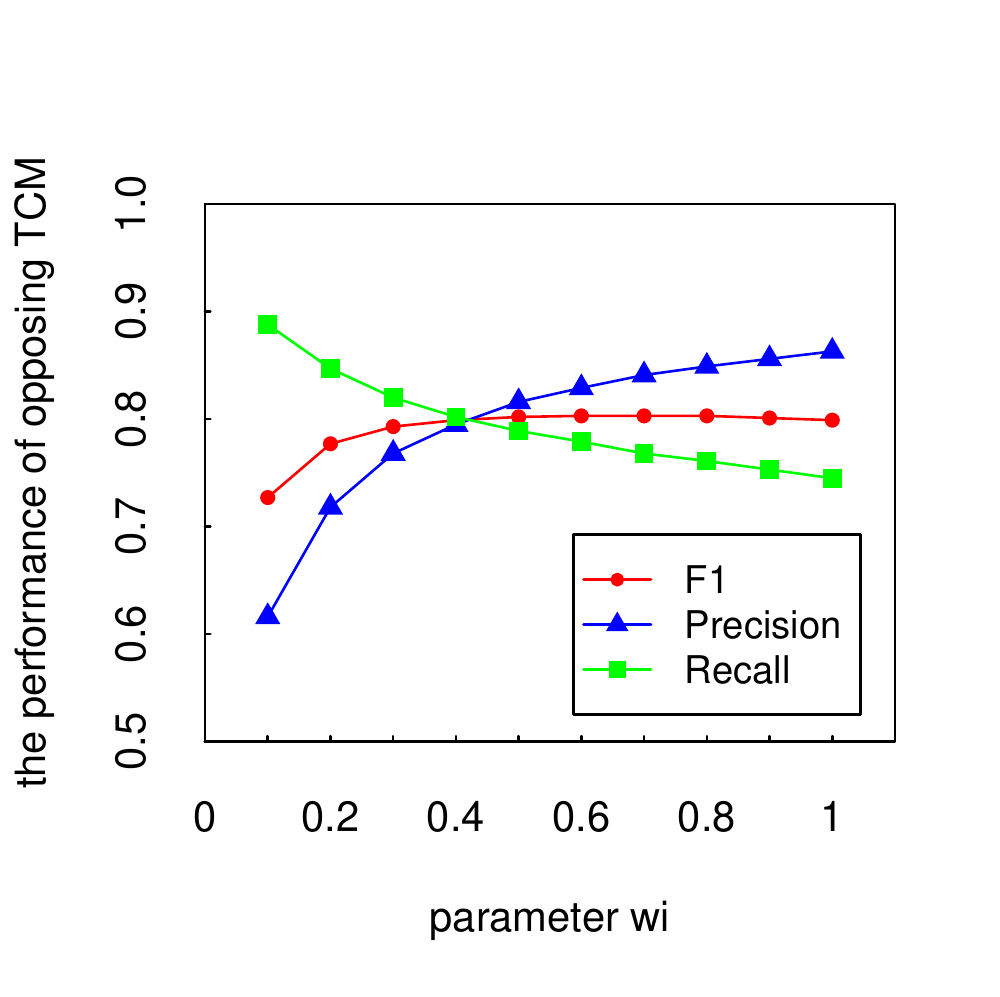}
     \label{}
     }
 \caption{
      {the Performance of Supporting TCM and Opposing TCM Separately.} 
       }
 \label{Figure_4}
 \end{figure}

\subsection{Classification Results}
Because the dataset is unbalanced, we tune the wi parameter for SVM, where  $0 \leq wi \leq 1$.

Figure 2 shows the performance of F1, precision and recall by varying the parameter wi from 0.1 to 1.0. when wi increases from 0 to 1, precision, recall and F1 all increase significantly and reach plateau. 

Figure 3 shows precision and recall separately with each class by varying the parameter wi. It is interesting that precision shows a reverse trend of that of recall. When wi increases from 0 to 1, the precision of supporting TCM gradually decreases while the precision of opposing TCM rapidly increase. During the same process, recall of supporting TCM increases while recall of opposing TCM significantly decreases. 

Figure 4 shows the performance of each class separately. From these figures we can see that it is better to set wi to 0.9. It summarizes the performance of the classifier of supporting TCM and the classifier of opposing TCM. When wi gradually increases, for TCM proponents, Precision decreases from 98$\%$ to 96$\%$, Recall increases from 91$\%$ to 98$\%$, and F1 increases gradually to a plateau phase. For TCM opponents, Precision increases 62$\%$ to 86$\%$, Recall decreases from 89$\%$ to 75$\%$, and F1 increases gradually to a plateau phase.    
 
Either viewing the whole or the individual class, when wi increases from 0.1 to 1, F1 value increases gradually to a plateau phase. F1 value reaches the optimal when wi equals to 0.9. 
\subsection{Adjusted Classification Results}
As introduced in Section 4.3, we can adjust the classification results based on the principle that the same user should have consistent opinions for a certain topic. Figure 5 shows the performance by varying the parameter $\gamma$ from 0.5 to 1(and fixing wi=0.9). There is a noticeable decline of F1. when $\gamma$ is set to 0.5, our model achieves the best performance of F1, which is 97$\%$.

\begin{figure}[t!]
\centering
\includegraphics[width=5cm]{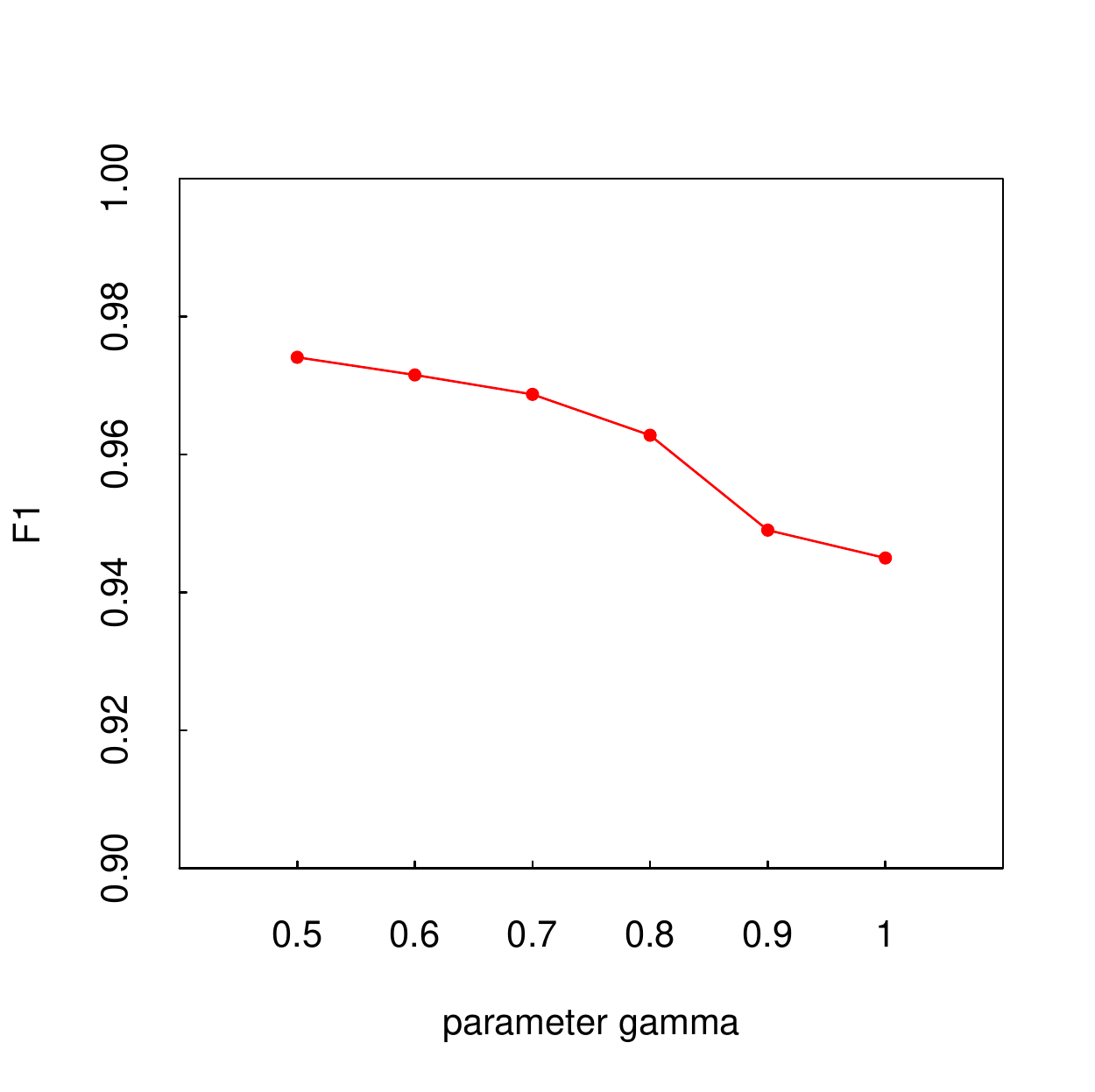}
\caption{
{\bf The Performance Curves  vs. Parameter $\gamma$.} 
}

\label{Figure_5}
\end{figure}
\subsection {Prediction}
Besides the labeled tweets, there are 1,602,634 unlabeled tweets which the topic is about TCM. We can predict their sentiment with our trained classifier. Figure 6 shows the curves for the number of tweets which, respectively, support TCM(a) and oppose TCM(b). The number of tweets supporting TCM far exceeds the number of tweets opposing TCM. For the simple comparison, the tweets number of both opposing and supporting TCM are converted to their log forms, as shown in (c). This result coincides with the real world. In china, most people support TCM, especially the regimen of TCM. There are only a small number of people opposing TCM. In addition, the tweet count before 2010 is very small, which is due to the limit of Weibo where only the most recent 2000 tweets of each user can be obtained. 

\begin{figure*}[t!]
\centering
\subfigure[SUPPORTING TCM]{
\includegraphics[width=5cm]{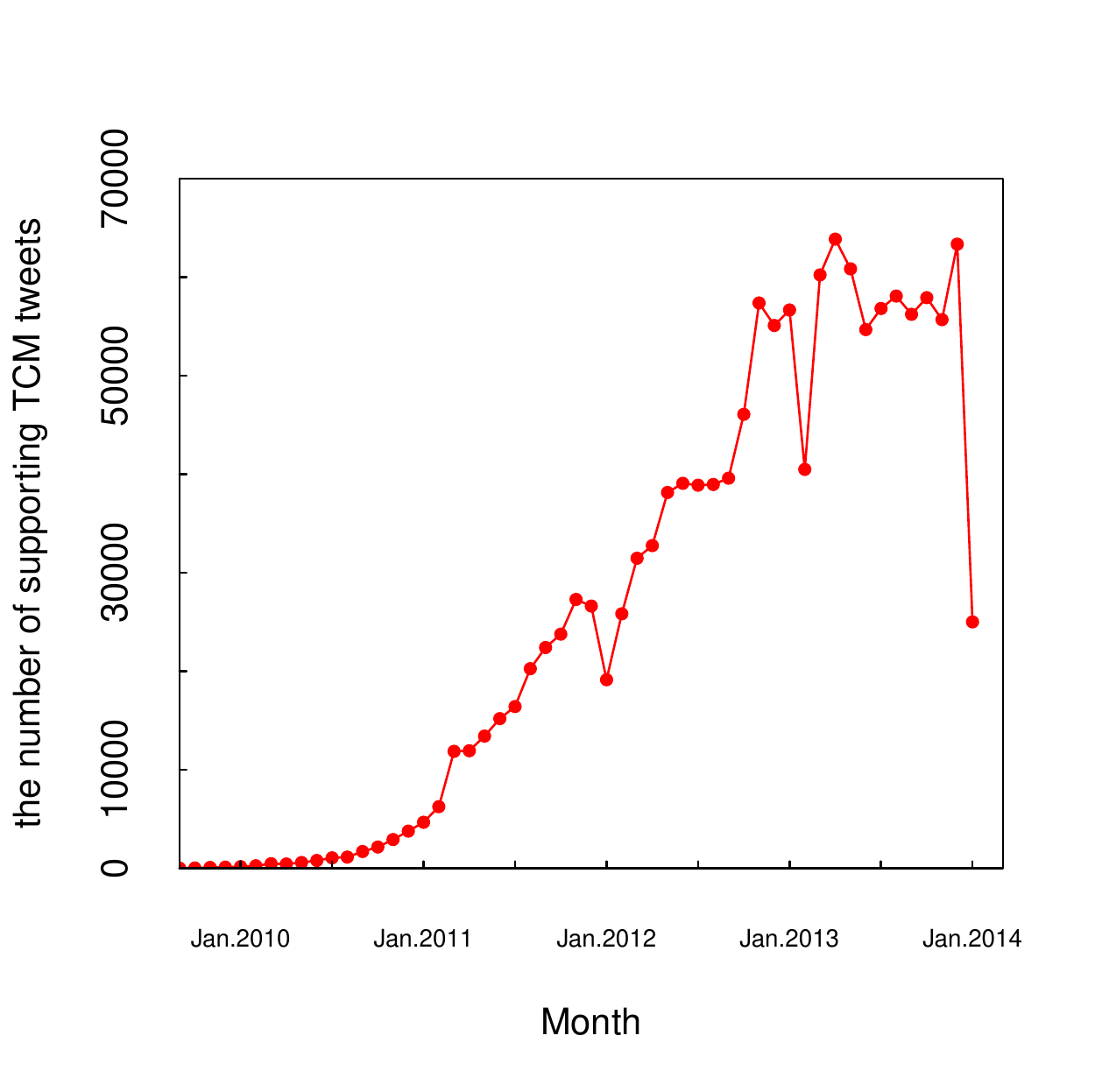}
\label{}
}
\subfigure[OPPOSING TCM]{
\includegraphics[width=5cm]{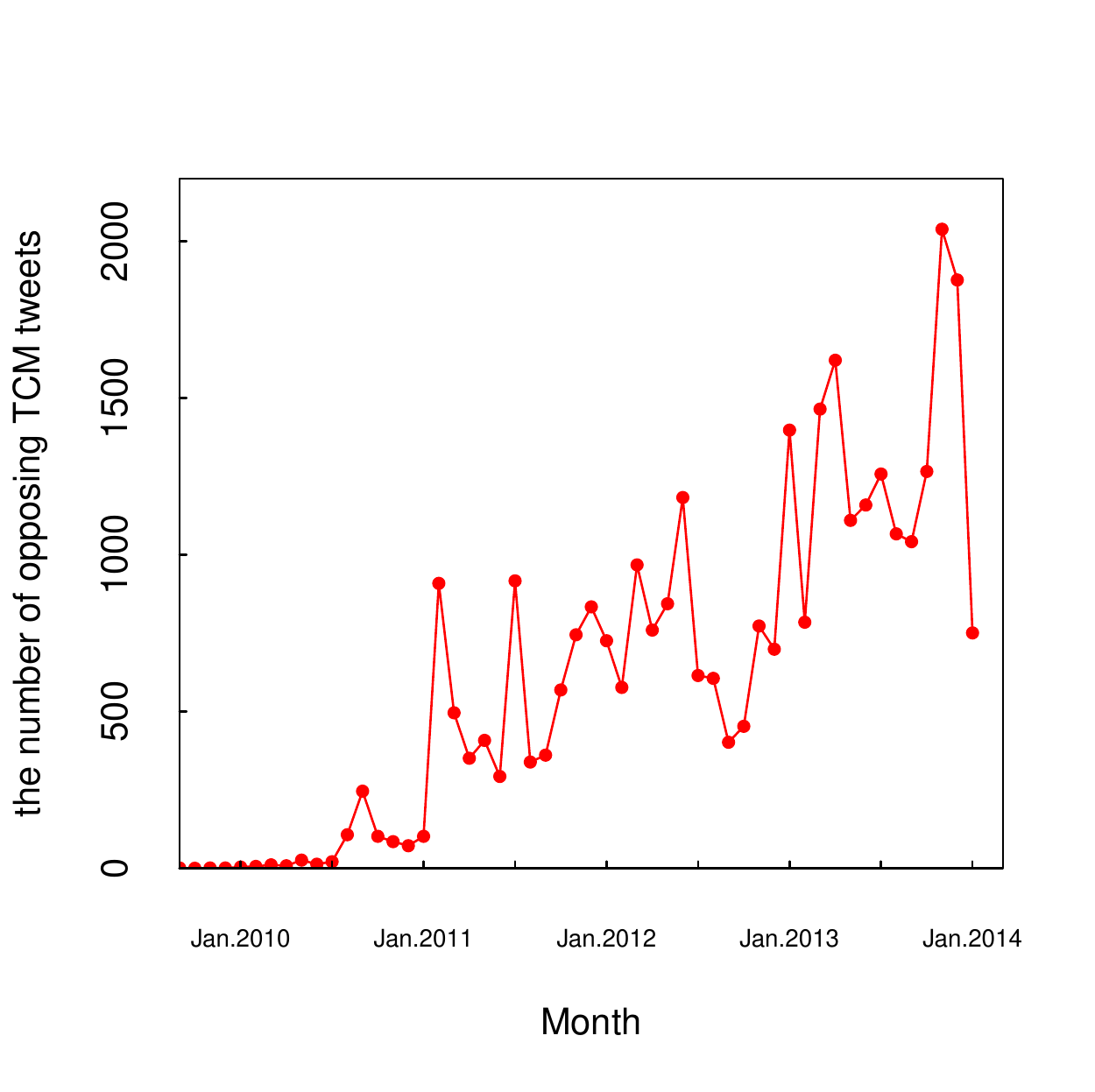}
\label{}
}
\subfigure[SUPPORTING TCM vs. OPPOSING TCM]{
\includegraphics[width=5cm]{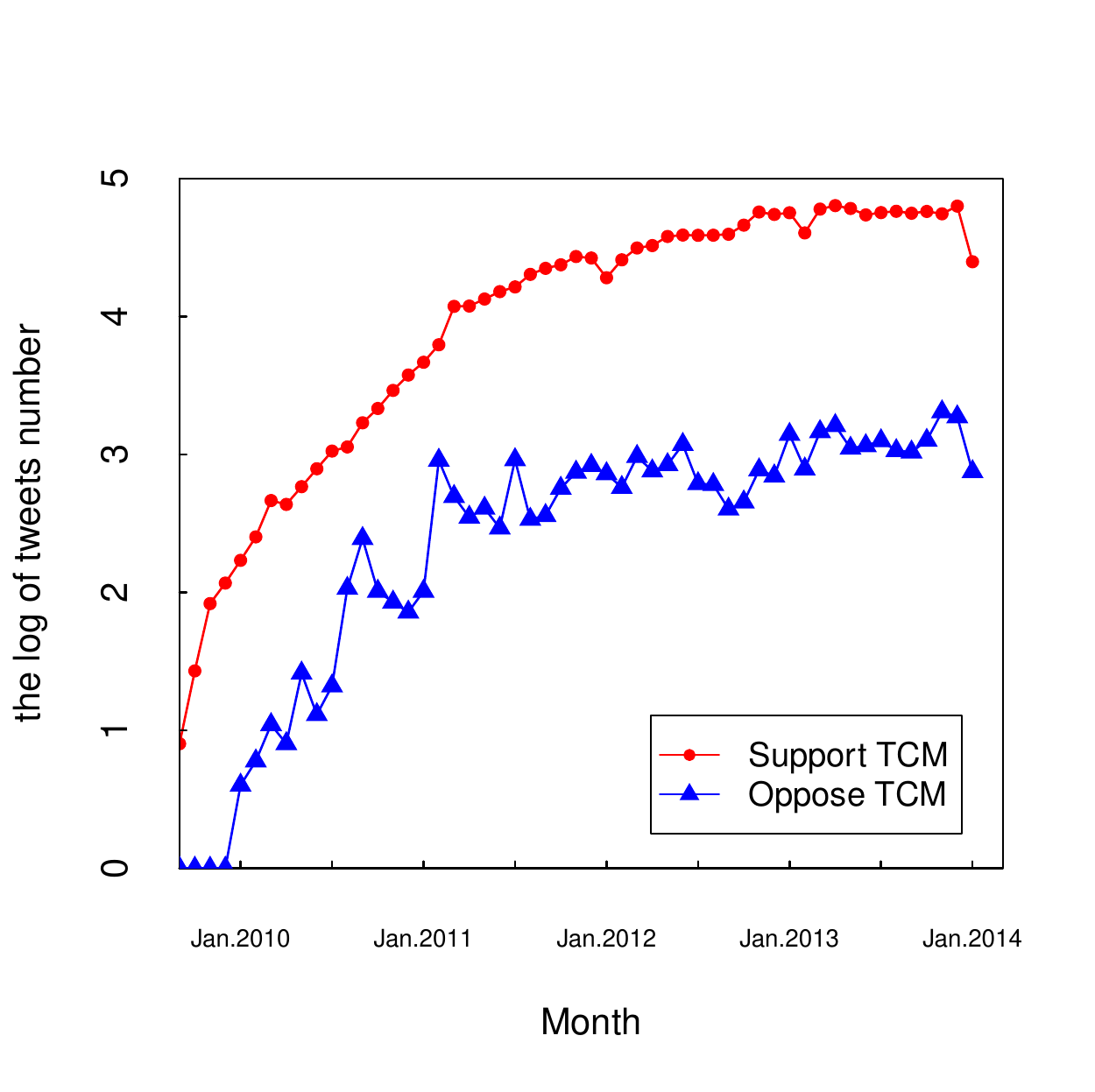}
\label{}
}
\caption{
{the Tweets Number of Supporting TCM vs. Opposing TCM.} 
}
\label{Figure_6}
\end{figure*}

After the sentiment classification of tweets concerning TCM, we can monitor the sentiment fluctuation of TCM in Weibo. As shown in Figure 6, the number of tweets supporting TCM decreases significantly during January of 2012, 2013 and 2014. Because the three periods coincide Chinese New Year, the decrease could be due to that people did not log on Weibo during these holiday seasons. On the contrary, number of tweets opposing TCM showed no clear trend. The erupt of the tweets opposing TCM could be caused by incidents related to TCM, which could be an interesting research topic in the future. 

 It is also found from Figure 6 that，the tweet counts of both class reached the peak in Nov. 2013. We show the details of the curve in that month in Figure 7. In November 2013, the number of tweets supporting TCM is relatively stable while the number of tweets opposing TCM fluctuates drastically. This is in line with overall trend of the number of each class.  

\begin{figure}[t!]
\centering
\includegraphics[width=5cm]{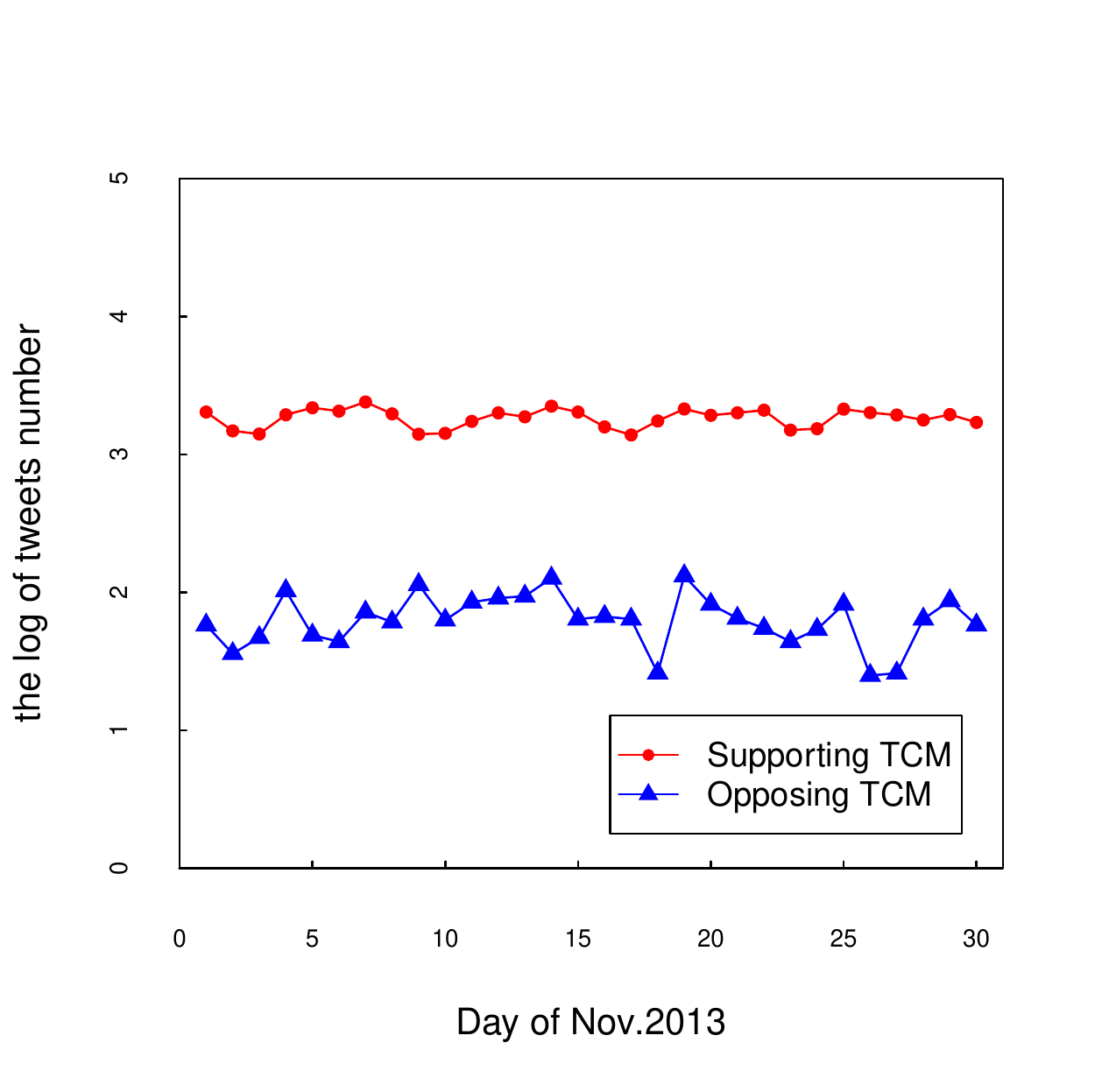}
\caption{
{\bf  the Prediction Result of Nov.2013.} 
}
\label{Figure_7}
\end{figure}

Moreover, Our methodology is able to generate keywords of TCM in favor tweets and TCM against tweets. The top 50 keywords of each class in Nov. 2013 is shown separately in Figure 8. "Traditional Chinese Medicine", "Health Preservation", "Food" etc. often appears in tweets supporting TCM, while "Chinese Patent Medicine", "injection", "toxicity" etc. frequently appears in tweets opposing TCM. This is conformed with Table 3. It is worth mentioning that words such as "toxic" or "harmful" appears in tweets supporting TCM too. This is not unexpected because TCM theory admits that a few TCM medicine is toxic so the dosage of these toxic TCM medicines should be controlled with caution. 

\begin{figure*}[t!]
\centering
\subfigure[SUPPORTING TCM]{
\includegraphics[width=7.5cm]{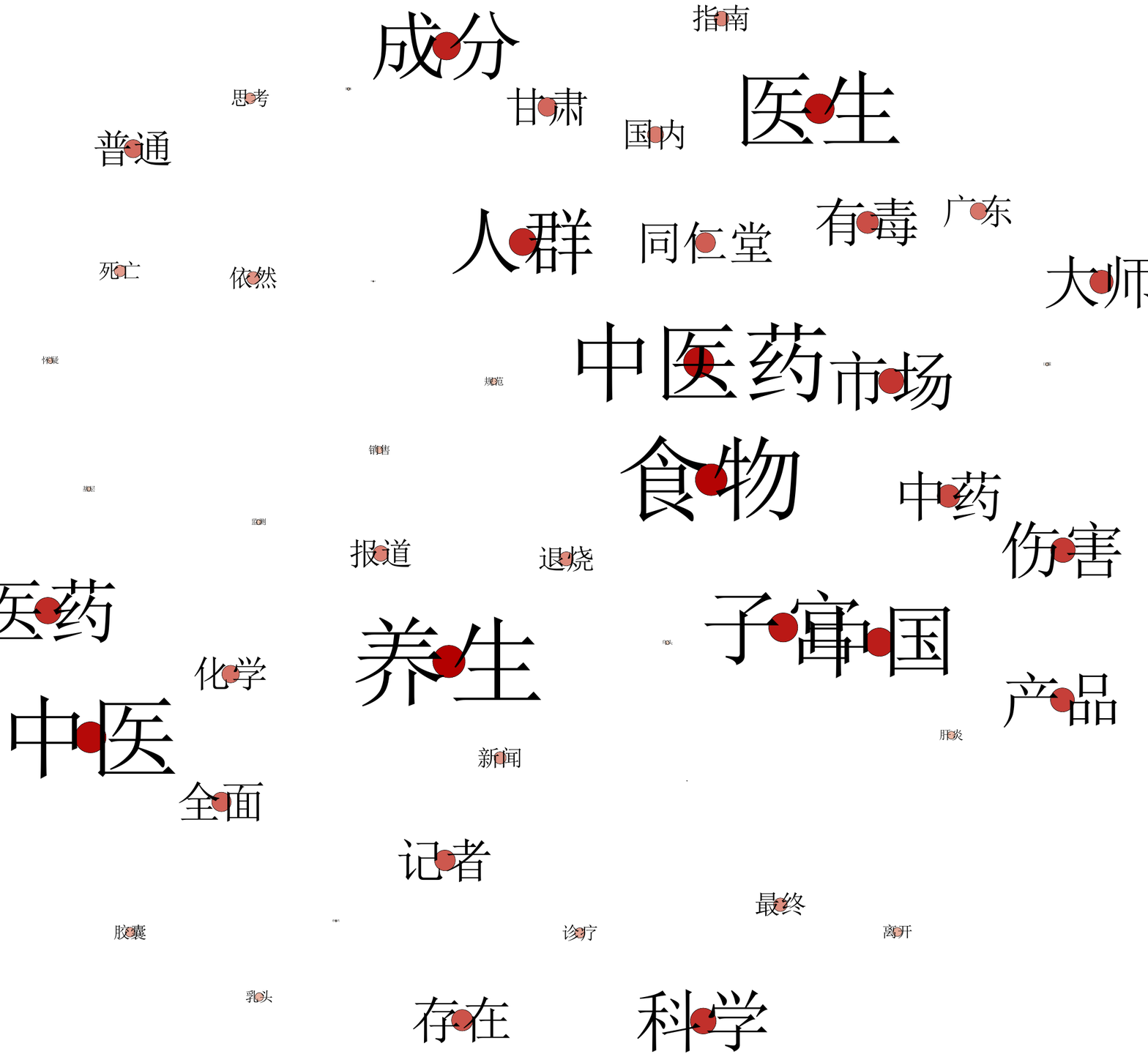}
\label{}
}
\subfigure[OPPOSING]{
\includegraphics[width=7.5cm]{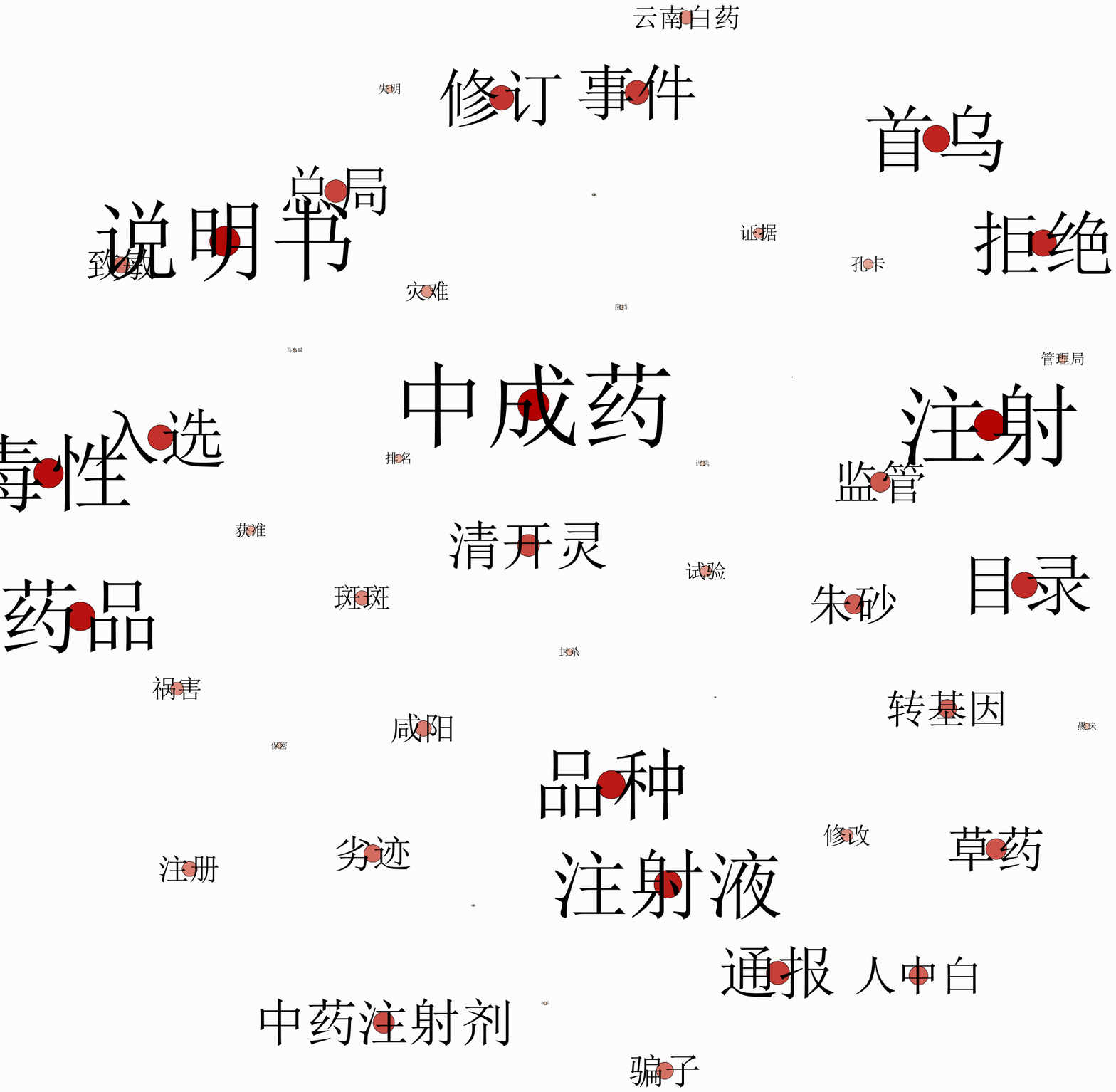}
\label{}
}
\caption{
{\bf  The Top 50 Key Words of Each Class in Nov.2013.} 
}

\label{Figure_8}
\end{figure*}

\section{Conclusion}
Traditional Chinese Medicine is an ancient but thriving and somewhat controversial discipline, meanwhile, it is important to study the public's sentiments and opinions on TCM. To the best of our knowledge, the present work is the first attempt to study sentiment analysis for TCM  based on user tag in Weibo. We classify the opinions on TCM into two categories: supporting TCM and opposing TCM. The F1 measure value of our method is 0.97. 

Moreover, we collect 48861 Weibo users who are interested in TCM and 1,650,497 tweets concerning TCM. And we construct two dictionary resources for processing Chinese tweets topic on TCM. Based on the aforementioned corpora and resources, we build an effective classifier with SVM to analyze the sentiment opinions on TCM using Weibo tweets automatically. 

In future work, we will explore more linguistic techniques to study sentiment analysis for TCM, such as parsing, semantic analysis and topic modeling.

{\bf Data sharing statement:} The unpublished data from this
study are available by contacting Junhui Shen (email:shen-junhui@163.com ; telephone: 86-10-64287566). Data can be sent by email.





%
\bibliographystyle{IEEEtran}
\bibliography{IEEEabrv,IEEEexample,tex}


\end{document}